\documentclass{article}
\usepackage{iclr2027_conference,times}
\usepackage{amsmath,amssymb,booktabs,graphicx,array}
\usepackage{subcaption}
\usepackage{multirow,threeparttable}
\usepackage{needspace}
\usepackage{flafter}
\usepackage{algorithm,algpseudocode}
\usepackage{fvextra}
\usepackage{tikz}
\usetikzlibrary{arrows.meta,positioning}
\usepackage{hyperref,xurl}
\newcommand{\system}{\textsc{Cairn}}
\newcommand{\fact}{\textsc{fact}}
\newcommand{\intent}{\textsc{intent}}
\DeclareRobustCommand{\loopbadge}[1]{\tikz[baseline=(badge.base)]
  \node[draw=black!55,circle,line width=.4pt,minimum size=8pt,
    inner sep=0pt,outer sep=0pt,font=\fontsize{5.5}{5.5}\selectfont,
    text height=4pt,text depth=0pt] (badge) {#1};}

\hypersetup{hidelinks,pdftitle={CAIRN: Dynamic Fact-Intent DAGs for Multi-Agent Exploration},pdfauthor={Zuyao Xu; Yuyang Jia; Junwei Guan; Xiang Li; KaiWen Shen; Zhiqiang Dong},pdfkeywords={LLM agents; Multi-agent coordination; Autonomous systems; Parallel exploration; Dependency graphs; Shared memory}}

\def\cairnSoftFigureStyle{1}
\def\cairnPanelFrames{1}
\title{CAIRN: Dynamic \textsc{Fact}--\textsc{Intent} DAGs for Multi-Agent Exploration}
\author{%
\makebox[\dimexpr\textwidth-2\tabcolsep\relax][l]{%
\textbf{Zuyao Xu\textsuperscript{\ensuremath{\dagger}\ensuremath{\|}}}\hfill
\textbf{Yuyang Jia\textsuperscript{\ensuremath{\ddagger}\ensuremath{\|}}}\hfill
\textbf{Junwei Guan\textsuperscript{\ensuremath{\dagger}}}\hfill
\textbf{Xiang Li\textsuperscript{\ensuremath{\dagger}*}}\hfill
\textbf{Kaiwen Shen\textsuperscript{\S}}\hfill
\textbf{Zhiqiang Dong\textsuperscript{\ensuremath{\ddagger}}}
}\\[4pt]
{\normalfont\small
\textsuperscript{\ensuremath{\dagger}} Nankai University\quad
\textsuperscript{\ensuremath{\ddagger}} Tencent Security Yunding Lab\quad
\textsuperscript{\S} Tsinghua University}\\[2pt]
{\normalfont\small
\href{mailto:xuzuyao@mail.nankai.edu.cn}{xuzuyao@mail.nankai.edu.cn}\quad
\href{mailto:lixiang@nankai.edu.cn}{lixiang@nankai.edu.cn}}%
}
\iclrfinalcopy 
\begin{document}
\maketitle
\begingroup
\renewcommand{\thefootnote}{\fnsymbol{footnote}}
\footnotetext[1]{Corresponding author}
\footnotetext[6]{Equal Contribution}
\endgroup
\begin{abstract}
LLM-powered autonomous systems have demonstrated promising capabilities in mathematical reasoning, engineering, and cybersecurity.
Yet how to organize these systems for effective, reliable, and sustained performance remains an open question.
In this paper, we present \system{}, a \textsc{fact}--\textsc{intent}-driven multi-agent paradigm for goal-directed exploration.
\system{} represents observations and planned investigations as a dynamic directed acyclic graph (DAG).
A \textsc{reasoner} interprets \textsc{fact}s to propose \textsc{intent}s, which \textsc{worker}s execute to produce new \textsc{fact}s.
Each \textsc{intent} references its supporting \textsc{fact}s and defines a potential exploration branch.
The persistent graph preserves goals, dependencies and findings across \textsc{worker}s, supporting knowledge reuse and parallel exploration.
The graph also makes execution trajectories traceable and auditable, providing a basis for human verification and intervention.
We evaluate \system{} across cybersecurity and mathematical reasoning tasks, examining task success, time to solution, and token consumption.
DAG-based coordination can incur higher token costs with no observable performance gains on tasks that require little effort.
However, on high-effort tasks ($\geq1$M tokens), we observe faster solutions in 76.5\% of cases, with speedups of up to $3.08\times$.
Moreover, as task effort increases, these time gains become more pronounced while relative token overhead declines, highlighting the potential of DAG-guided parallel exploration.
\end{abstract}

\section{Introduction}
\label{sec:introduction}

\begin{figure}[!ht]
\centering
\resizebox{\linewidth}{!}{\input{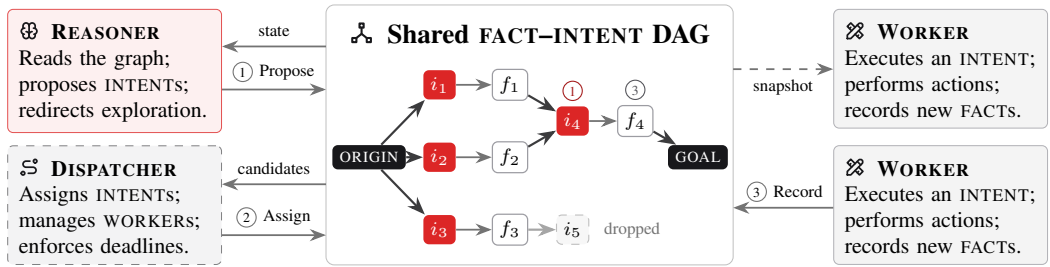}}
\caption{Exploration over a shared \textsc{fact}--\textsc{intent} DAG.
The \textsc{reasoner} reads graph state and proposes \textsc{intent}s citing source \textsc{fact}s (\loopbadge{1} Propose).
The dashed \textsc{dispatcher} receives candidate \textsc{intent}s and assigns eligible candidates (\loopbadge{2} Assign).
\textsc{worker}s execute assigned \textsc{intent}s using a graph snapshot and record outcome \textsc{fact}s (\loopbadge{3} Record).
In the illustrated graph, $f_1$ and $f_2$ support $i_4$.
The \textsc{reasoner} drops $i_5$ when new evidence contradicts its underlying premise.}
\label{fig:loop}
\end{figure}

Many difficult intellectual tasks require extended effort and collaboration for humans to solve, yet recent successes of large language models (LLMs) suggest that multi-agent systems can achieve similar results in a fraction of the time~\citep{openai2026navierstokes,anthropic2026glasswing}.
For example, coordinated agents have contributed to research on two Millennium Prize Problems, with a reported breakthrough on the Navier--Stokes problem and new results related to the Riemann hypothesis~\citep{openai2026navierstokes,anthropic2026zeta}.
Project Glasswing and OpenAI's Daybreak likewise incorporate multi-agent workflows into security efforts, with Glasswing reporting over 10,000 high- or critical-severity vulnerabilities found in one month and Daybreak reporting 143 upstream-accepted patches~\citep{anthropic2026glasswing,openai2026daybreak,openai2026deepsecurityscan}.
Together, these advances point to a new level of capability in harnessing LLMs through multi-agent collaboration.

\begin{figure}[!t]
\centering
\resizebox{\linewidth}{!}{\begin{tikzpicture}[>=Stealth,font=\fontsize{8}{9.4}\selectfont,
  panel/.style={draw=black!45,rounded corners=3pt,line width=.6pt},
  box/.style={draw=black!25,rounded corners=2pt,align=left,inner sep=4pt},
  reasoning/.style={box,fill=black!4},
  intent/.style={box,fill=orange!12,draw=orange!50!black},
  fact/.style={box,fill=green!12,draw=green!40!black},
  boundary/.style={box,fill=blue!12,draw=blue!50!black},
  flow/.style={->,draw=black!55,line width=.65pt},
  dep/.style={->,draw=green!35!black,line width=.8pt}]
\ifdefined\cairnSoftFigureStyle
  \input{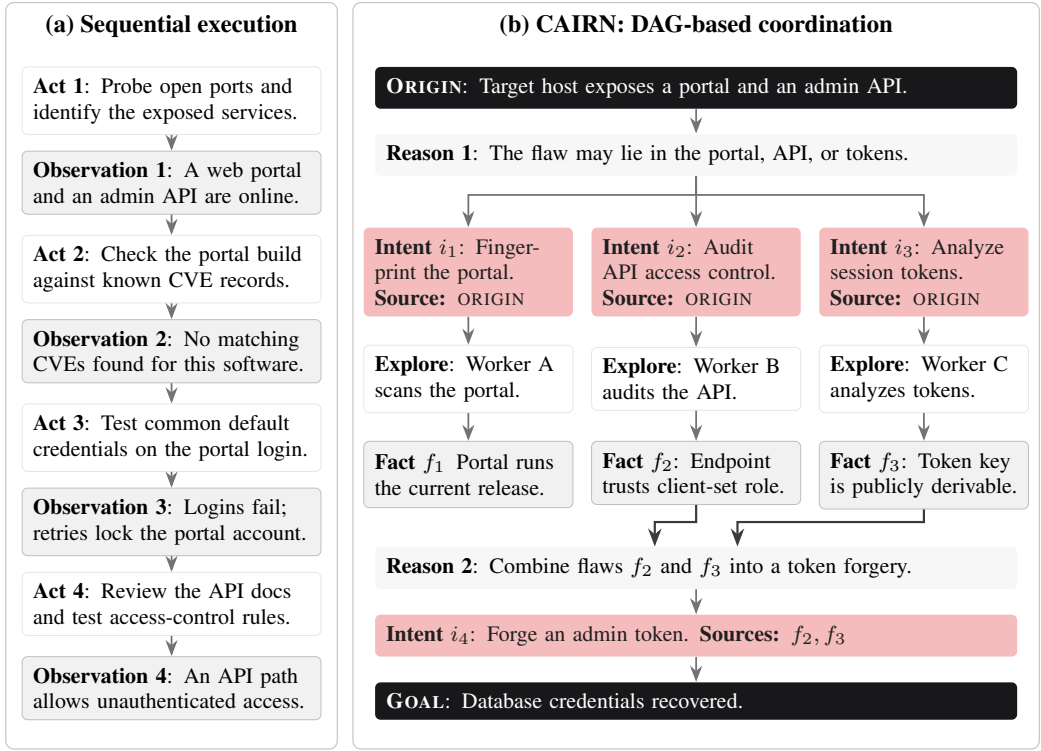}
  \tikzset{
    panel/.style={draw=black!22,rounded corners=3pt,line width=.45pt},
    box/.style={draw=black!10,rounded corners=2pt,line width=.3pt,
      align=left,inner sep=4pt},
    reasoning/.style={box,fill=black!3,draw=none},
    intent/.style={box,fill=cairnTrialIntent!31,
      draw=none,text=black!88},
    fact/.style={box,fill=cairnTrialFact!10,draw=cairnTrialFact!45},
    boundary/.style={box,fill=cairnTrialBoundary,draw=none,text=white},
    dep/.style={->,draw=cairnTrialBoundary!85,line width=.8pt}}
\fi
\path[use as bounding box] (-.1,.90) rectangle (13.35,-9.81);
\node[box,text width=12.8cm,fill=black!2,anchor=north west] at (0,.85)
  {\textbf{Illustrative task:} Investigate a host exposing a web portal and an admin API protected by session tokens.\\
  The portal stores its data in a backend database. Recover the credentials needed to access that database.};

\draw[panel] (0,-.18) rectangle (4.25,-9.75);
\draw[panel] (4.45,-.18) rectangle (13.25,-9.75);
\node[font=\bfseries\fontsize{9}{11}\selectfont] at (2.125,-.48)
  {(a) Sequential execution};
\node[font=\bfseries\fontsize{9}{11}\selectfont] at (8.85,-.48)
  {(b) CAIRN: DAG-based coordination};

\node[box,text width=3.55cm,anchor=north] (la1) at (2.125,-1.00)
  {\textbf{Act 1}: Probe open ports and\\identify the exposed services.};
\node[fact,text width=3.55cm,anchor=north] (lo1) at (2.125,-2.08)
  {\textbf{Observation 1}: A web portal\\and an admin API are online.};
\node[box,text width=3.55cm,anchor=north] (la2) at (2.125,-3.16)
  {\textbf{Act 2}: Check the portal build\\against known CVE records.};
\node[fact,text width=3.55cm,anchor=north] (lo2) at (2.125,-4.24)
  {\textbf{Observation 2}: No matching\\CVEs found for this software.};
\node[box,text width=3.55cm,anchor=north] (la3) at (2.125,-5.32)
  {\textbf{Act 3}: Test common default\\credentials on the portal login.};
\node[fact,text width=3.55cm,anchor=north] (lo3) at (2.125,-6.40)
  {\textbf{Observation 3}: Logins fail;\\retries lock the portal account.};
\node[box,text width=3.55cm,anchor=north] (la4) at (2.125,-7.48)
  {\textbf{Act 4}: Review the API docs\\and test access-control rules.};
\node[fact,text width=3.55cm,anchor=north] (lo4) at (2.125,-8.56)
  {\textbf{Observation 4}: An API path\\allows unauthenticated access.};
\foreach \a/\b in {la1/lo1,lo1/la2,la2/lo2,lo2/la3,la3/lo3,lo3/la4,la4/lo4}
  \draw[flow] (\a.south)--(\b.north);

\node[boundary,text width=7.95cm,anchor=north] (f0) at (8.85,-1.00)
  {\textbf{\textsc{Origin}}: Target host exposes a portal and an admin API.};
\node[reasoning,text width=7.95cm,anchor=north] (r1) at (8.85,-1.86)
  {\textbf{Reason 1}: The flaw may lie in the portal, API, or tokens.};
\node[intent,text width=2.4cm,anchor=north] (i1) at (5.93,-3.06)
  {\textbf{Intent $i_1$}: Fingerprint the portal.\\\textbf{Source:} \textsc{origin}};
\node[intent,text width=2.4cm,anchor=north] (i2) at (8.85,-3.06)
  {\textbf{Intent $i_2$}: Audit API access control.\\\textbf{Source:} \textsc{origin}};
\node[intent,text width=2.4cm,anchor=north] (i3) at (11.77,-3.06)
  {\textbf{Intent $i_3$}: Analyze session tokens.\\\textbf{Source:} \textsc{origin}};
\node[fact,text width=2.4cm,minimum height=.85cm,anchor=north] (f1) at (5.93,-5.80)
  {\textbf{Fact $f_1$}\enspace Portal runs\\the current release.};
\node[fact,text width=2.4cm,anchor=north] (f2) at (8.85,-5.80)
  {\textbf{Fact $f_2$}: Endpoint trusts client-set role.};
\node[fact,text width=2.4cm,anchor=north] (f3) at (11.77,-5.80)
  {\textbf{Fact $f_3$}: Token key is publicly derivable.};
\path (i1.south)--(f1.north)
  node[midway,box,text width=2.4cm] (e1)
  {\textbf{Explore}: Worker A scans the portal.};
\path (i2.south)--(f2.north)
  node[midway,box,text width=2.4cm] (e2)
  {\textbf{Explore}: Worker B audits the API.};
\path (i3.south)--(f3.north)
  node[midway,box,text width=2.4cm] (e3)
  {\textbf{Explore}: Worker C analyzes tokens.};
\node[boundary,text width=7.95cm,anchor=south] (f4) at (f0.center |- lo4.south)
  {\textbf{\textsc{Goal}}: Database credentials recovered.};
\node[intent,text width=7.95cm,anchor=south,yshift=.32cm] (i4) at (f4.north)
  {\textbf{Intent $i_4$}: Forge an admin token. \textbf{Sources:} $f_2,f_3$};
\node[reasoning,text width=7.95cm,anchor=south,yshift=.32cm] (r2) at (i4.north)
  {\textbf{Reason 2}: Combine flaws $f_2$ and $f_3$ into a token forgery.};
\draw[flow] (f0)--(r1);
\draw[flow] (r1.south)--++(0,-.24)-|(i1.north);
\draw[flow] (r1.south)--++(0,-.24)-|(i2.north);
\draw[flow] (r1.south)--++(0,-.24)-|(i3.north);
\foreach \a/\b in {i1/e1,e1/f1,i2/e2,e2/f2,i3/e3,e3/f3,r2/i4}
  \draw[flow] (\a.south)--(\b.north);
\draw[flow] (i4.south)--(f4.north);
\draw[dep] (f2.south)--(8.85,-6.84)-|([xshift=-.53cm]r2.north);
\draw[dep] (f3.south)--(11.77,-6.84)-|([xshift=.53cm]r2.north);
\end{tikzpicture}}
\caption{Two approaches to the same credential-recovery task.
(a) A sequential action--observation trajectory.
(b) \system{}'s \textsc{fact}--\textsc{intent} DAG. Three investigations start from the \textsc{origin} and examine the portal, API, and session tokens. The API and token findings jointly motivate a subsequent investigation toward the \textsc{goal}.}
\label{fig:dag}
\end{figure}

Despite these successes, a central question remains: \textit{how to coordinate multiple agents to collaborate effectively and sustainably?}
Prior work offers several approaches.
MetaGPT and ChatDev organize software-development agents through specialized roles and workflows~\citep{hong2024metagpt,qian2024chatdev}.
AutoGen provides a programmable framework for multi-agent conversation~\citep{wu2024autogen}.
More recent frameworks dynamically create task-specific sub-agents~\citep{ruan2026aorchestra} and coordinate dependent subtasks through verification-aware plans~\citep{xu2026verimap}.

In this paper, we focus on \emph{heuristic problem solving}: tasks with a clear objective and an open solution path.
Solving such tasks may require pursuing multiple possible directions over time.
Candidate approaches may be proposed, tested, refined, or abandoned as understanding develops.
Motivated by such problems, we introduce \system{}, a multi-agent coordination paradigm that represents findings as \textsc{fact}s and planned investigations as \textsc{intent}s.
A \textsc{reasoner} proposes \textsc{intent}s that explicitly reference their supporting \textsc{fact}s, and \textsc{worker}s execute these \textsc{intent}s, contributing new \textsc{fact}s to the graph (Figure~\ref{fig:loop}).
These dependencies form a persistent directed acyclic graph: one \textsc{fact} can motivate several \textsc{intent}s, and several \textsc{fact}s can jointly motivate a new \textsc{intent}.
The graph preserves these connections across \textsc{worker}s and sessions, helping agents reuse findings and coordinate parallel exploration.
It also makes each investigation traceable, showing what motivated it, what it found, and how findings from different branches informed later directions.

We evaluate \system{} on twelve public mathematical reasoning problems from FrontierMath~\citep{epoch2026frontiermath} and forty cybersecurity tasks from Cybench~\citep{zhang2025cybench}, comparing it with a baseline agent using the same model, agent loop, and task environment.
On high-effort tasks ($\geq1$M baseline tokens), \system{} finishes sooner in 76.5\% of cases, with speedups of up to $3.08\times$.
Although \system{} uses more tokens on average, its relative token overhead decreases as task effort increases.
We also manually reviewed sixteen \fact{}--\intent{} DAGs from FrontierMath runs, finding that they reveal how investigations reuse earlier findings, combine evidence, and correct errors.
These graphs provide structured records for human review, verification, and auditing of multi-agent problem solving.
Together, our results suggest a role for coordinated exploration in sustained tasks where faster completion justifies additional computation, including hyperscale deployments with abundant inference capacity.
In practice, an open-source implementation of this paradigm has attracted approximately 3,000 GitHub stars~\citep{cairn2026github}.
A system built on the paradigm was the only one to solve all tasks in the second Tencent Cloud Intelligent Penetration Hackathon~\citep{tsechackathon2026official} and achieved a leading result in the public TSecBench cybersecurity evaluation~\citep{tsecbench2026public}.

\section{CAIRN: Exploration through \fact{}--\intent{} DAGs}
\label{sec:design}

Figure~\ref{fig:dag} presents two approaches to the same credential-recovery task: a sequential action--observation trajectory in panel~(a) and \system{}'s \fact{}--\intent{} DAG in panel~(b).
In panel~(b), \system{} represents the task through an \textsc{origin}, which records the initial information, and a \textsc{goal}, which specifies the desired outcome.
\system{} records planned investigations as \intent{}s and their findings as \fact{}s, linking each \intent{} to the \fact{}s that motivate it and the findings it produces.
Task solving is thus represented as the incremental construction of a persistent directed acyclic graph (DAG) from the \textsc{origin} toward the \textsc{goal}.

\subsection{Task solving as graph construction}
\system{} constructs the task graph as it investigates possible routes from the \textsc{origin} toward the \textsc{goal}.
Proposing an investigation adds an \intent{} linked to one or more existing \fact{}s that motivate it.
As the investigation proceeds, its findings are added as new \fact{}s linked to that \intent{}.
These findings can motivate further investigations, extending the graph through successive \fact{}--\intent{}--\fact{} steps.
When the accumulated evidence is judged to satisfy the task objective, a completion \intent{} connects the supporting \fact{}s to the terminal \textsc{goal}.

Starting with the \textsc{origin} and \textsc{goal} nodes, exploration extends the graph through three operations:
\begin{enumerate}
\item \textit{Propose:} add an \intent{} citing the \fact{}s that motivate it.
\item \textit{Assign:} dispatch an eligible \intent{} to a \textsc{worker} for investigation.
\item \textit{Record:} publish findings as new \fact{}s that can support later \intent{}s.
\end{enumerate}

The graph supports collaboration through \emph{branching} and \emph{convergence}.
Shared \fact{}s can motivate separate \intent{}s, allowing agents to pursue different directions; findings from these branches can then jointly support a subsequent \intent{}.
A formal definition of the graph and its operations is provided in Appendix~\ref{app:algorithms}.

\begin{figure}[!t]
\centering
\resizebox{.94\linewidth}{!}{\input{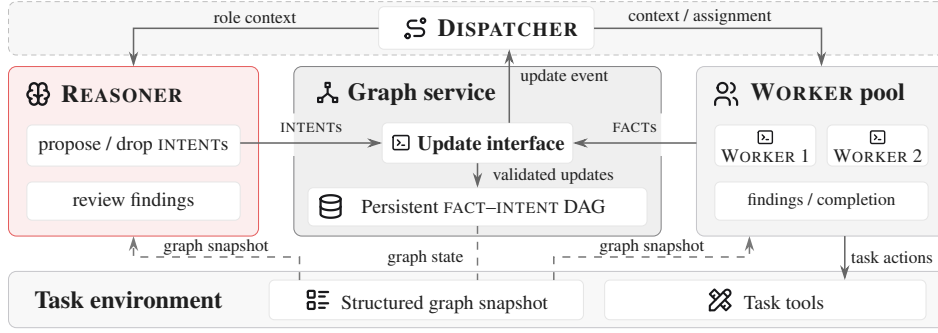}}
\caption{\system{} execution architecture.
The \textsc{dispatcher} starts \textsc{reasoner} and \textsc{worker} sessions with graph snapshots.
\textsc{Sessions} submit changes through the graph service; \textsc{worker}s also use task tools.
Accepted updates persist for later sessions and notify the \textsc{dispatcher}.
Solid arrows show assignments, updates, and task actions; dashed links show graph state and snapshots.}
\label{fig:implementation}
\end{figure}

\subsection{Coordination through a shared graph}
\label{sec:shared-graph}

We implement \system{} with two agent roles: a \textsc{reasoner} and a set of \textsc{worker}s (Figure~\ref{fig:implementation}).
The DAG serves as a persistent shared memory, making recorded evidence and investigation state available across agents.
The \textsc{reasoner} uses this memory to propose \intent{}s and redirect investigations, while \textsc{worker}s carry out assigned investigations using task tools and publish findings as \fact{}s.
A rule-based \textsc{dispatcher} initiates reasoning and assigns eligible \intent{}s to \textsc{worker}s.
Each invocation of either agent role constitutes a \emph{session}, initialized with a graph snapshot and role-specific context.
Both roles submit updates through a graph service that validates and persists accepted changes.

Independent investigations can proceed in parallel.
Once accepted, a finding is available to later sessions while its producing \textsc{worker} continues the investigation.
The \textsc{reasoner} can use these findings to propose follow-up investigations, allowing branches to develop alongside ongoing work.

\subsection{Inspectable trajectories and revision}
The graph's source and output edges expose an inspectable investigation trajectory: which findings motivated a direction, what it produced, and where branches split or join.
The graph traces investigations and their relationships at a higher level than individual individual agent actions, 
A \fact{} records an agent-reported finding whose reliability may change as evidence accumulates; it may describe a failed attempt, a provisional observation, or a claim later contradicted by new evidence.

For robustness, the \textsc{reasoner} have the ability to check and drop \textsc{intent}s, stopping further updates from any assigned \textsc{worker}s, allowing it to correct earlier mistakes or redirect investigations,.
A \textsc{worker} can publish corrective evidence as a new \fact{}, which replacement \intent{}s can cite to extend the recorded trajectory with new branches.
The graph preserves the investigation history, including the records of dropped \intent{}s, allowing later reviewers to understand the reasoning behind the revision.
We present a detailed description of \system{}'s implementation in Appendix~\ref{app:protocol}.

\section{Experiment Setup}
\label{sec:experimental-setup}
\begin{figure}[!t]
\centering
\resizebox{\linewidth}{!}{\input{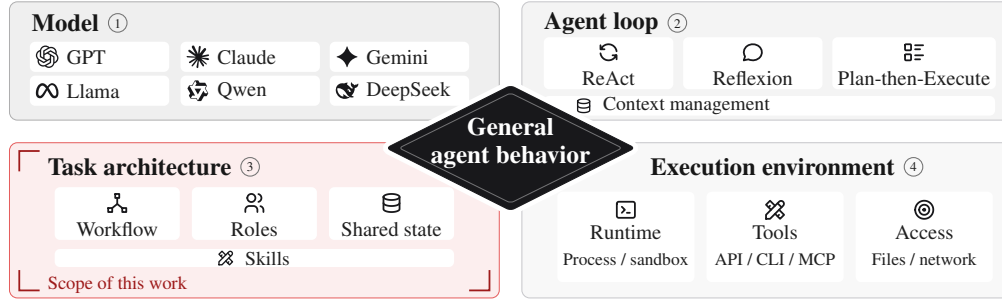}}
\caption{General agent behavior design factors.
\system{} focuses on task architecture as a multi-agent coordination paradigm, while other factors are held constant as the basis for comparison in the evaluation.
The examples in each area illustrate representative design choices.}
\label{fig:agent-design-factors}
\end{figure}

In this section, we describe the experimental setup for evaluating \system{}.
We summarize four factors shaping agent behavior in Figure~\ref{fig:agent-design-factors}.
The \emph{model} affects reasoning capabilities and learned knowledge, while the \emph{agent loop} shapes tool interaction and context management.
The \emph{task architecture} governs how work is organized and how information is retained and shared.
The \emph{execution environment} determines the runtime facilities, task tools, and access permissions during action.

Among these factors, \system{} is positioned at the task-architecture level, coordinating multi-agent exploration through a shared \fact{}--\intent{} graph.
To provide a transparent basis for assessing its effect on task solving, we compare \system{} with a direct agent baseline (referred to as \textsc{Direct}) using the same model, agent-loop implementation, and task environment, treating the high-level coordination paradigm as the variable of interest.

\paragraph{Benchmarks.}
We choose two benchmarks with externally checkable outcomes and a range of task durations, both used in prior work~\citep{udeshi2025dcipher,kouremetis2026cheating,zheng2026comath,mcfadyen2026compute}:
\begin{enumerate}
\item \textit{FrontierMath} is a mathematical reasoning benchmark with problems of increasing difficulty from Tier~1 to Tier~4.
According to Epoch AI, Tier~3 and Tier~4 problems can occupy specialists for days; leading mathematicians have described them as ``exceptionally challenging''~\citep{epoch2026frontiermath}.
We use the twelve publicly available sample problems.

\item \textit{Cybench} is a cybersecurity benchmark that provides CTF tasks with interactive environments and externally checkable outcomes \citep{zhang2025cybench}.
We use the unguided complete-task setting: solvers receive the original task, permitted files, and target access without benchmark subtask hints.
\end{enumerate}

\paragraph{Systems setup.}
We use Pi~0.85.1~\citep{earendil2025pi} as our minimal agent implementation and GLM-5.1~\citep{zai2026glm51} as the shared model.
\textsc{Direct} retains Pi's native system prompt, local tools, and context management; \system{} replaces the prompt with role-specific instructions and adds graph-based collaboration tools, with up to six execution slots shared by the \textsc{reasoner} and \textsc{worker}s.
Both systems run in isolated containers built from the same benchmark-specific image, with public Internet access disabled.
Within a five-hour hard limit, agents may resubmit after feedback from external verifiers.
We report solving time to the first accepted answer, assigning five hours to runs that reach the limit unsolved, and total token usage per run.
Configuration, environment, and measurement details appear in Appendix~\ref{app:manifest}.

\section{Results}
\label{sec:results}

In this section, we present the results of our evaluation of \system{} on two benchmarks: FrontierMath~\citep{epoch2026frontiermath} and Cybench~\citep{zhang2025cybench}.

\begin{table*}[t]
\centering
\footnotesize
\setlength{\tabcolsep}{1.6pt}
\renewcommand{\arraystretch}{1.12}
\captionsetup{position=top,skip=10pt,justification=justified,singlelinecheck=false,font={normalfont,normalsize},labelfont=normalfont,textfont=normalfont}
\makeatletter
\let\cairnInputTableRows\@@input
\makeatother
\begin{minipage}[t]{0.37\linewidth}
\vspace{0pt}
\captionsetup{type=table}
\caption{Benchmark-level summary of mean solving time and token usage on FrontierMath and Cybench.}
\label{tab:results-summary}
\begin{threeparttable}
\begin{tabular*}{\linewidth}{@{\extracolsep{\fill}}lrr@{}}
\toprule
Metric & \textsc{Direct} & \system{} \\
\midrule
\cairnInputTableRows tables/benchmark_summary_rows.tex
\bottomrule
\end{tabular*}
\begin{tablenotes}[para,flushleft]
\footnotesize
\setlength{\itemsep}{2pt}
\item[a] Runs: evaluation runs. 
\item[b] Faster: paired tasks solved sooner.
\end{tablenotes}
\end{threeparttable}

\end{minipage}\hfill
\begin{minipage}[t]{0.60\linewidth}
\vspace{0pt}
\captionsetup{type=table}
\caption{Time--compute profile by task effort.
Effort is \textsc{Direct}'s mean token usage for solving a task. For each group, we report run counts, mean solving time and token usage, and mean per-task ratios for \textsc{Direct} and \system{}.}
\label{tab:effort-summary}
\begin{threeparttable}
\begin{tabular*}{\linewidth}{@{\extracolsep{\fill}}llrrrrr@{}}
\toprule
Effort & Sys.\tnote{a} & Runs & Time & Tokens & \multicolumn{2}{c}{Mean ratio\tnote{b}} \\
(M), $n$ & & & (min) & (M) & Time$\uparrow$ & Tokens$\downarrow$ \\
\midrule
\cairnInputTableRows tables/pooled_effort_rows.tex
\bottomrule
\end{tabular*}
\begin{tablenotes}[para,flushleft]
\footnotesize
\setlength{\itemsep}{2pt}
\item[a] Sys. = system; D = direct agent; C = CAIRN; $n$ = tasks.
\item[b] Time ratio: D/C; token ratio: C/D.
$\uparrow$/$\downarrow$: higher/lower better.
\end{tablenotes}
\end{threeparttable}

\end{minipage}
\end{table*}

\subsection{Benchmark-level results}
\label{sec:benchmark-results}
Table~\ref{tab:results-summary} summarizes solving time and token usage on each benchmark.
On FrontierMath, \system{} has a lower mean solving time than \textsc{Direct} (27.45 versus 53.19 minutes) and finishes sooner on 8 of 12 problems (66.7\%), including all five Tier-3 and Tier-4 problems.
This time advantage comes with higher mean total token usage: 13.44M for \system{} versus 2.81M for \textsc{Direct}.
On Cybench, \textsc{Direct} finishes sooner on 30 of 40 tasks (75\%), with lower mean solving time (7.31 versus 8.49 minutes) and token usage (0.83M versus 3.22M).
The contrast between benchmarks motivates us to examine the results further at the task level.

\subsection{Task-dependent performance}
\label{sec:task-dependent-performance}
\begin{figure}[!t]
\centering
\includegraphics[width=\linewidth]{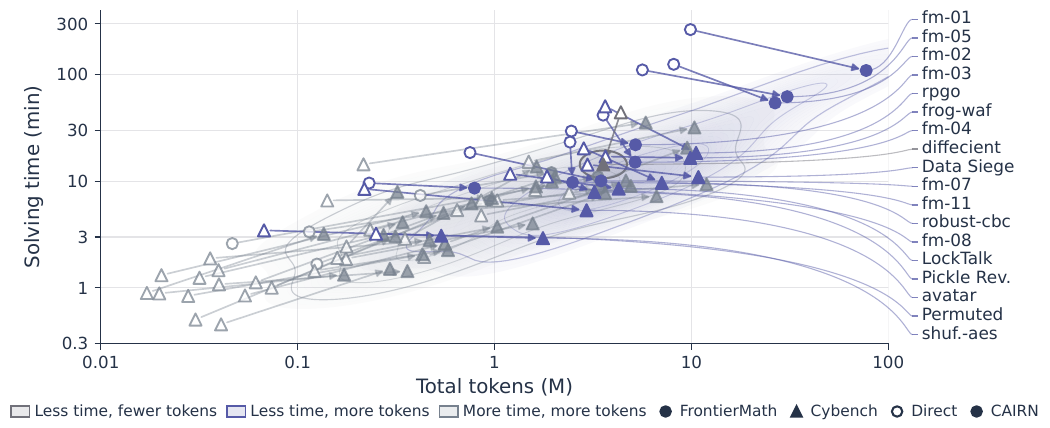}
\caption{Time--token trade-offs across the 52 tasks.
Arrows run from \textsc{Direct} to \system{};
labels on the right name the tasks on which \system{} finishes sooner than \textsc{Direct}; lines connect each label to its \system{} endpoint.
Shading shows where \system{} endpoints cluster within each outcome group.}
\label{fig:time-token-outcomes}
\end{figure}

\begin{figure}[!t]
\centering
\begin{subfigure}{0.48\linewidth}
\centering
\includegraphics[width=\linewidth]{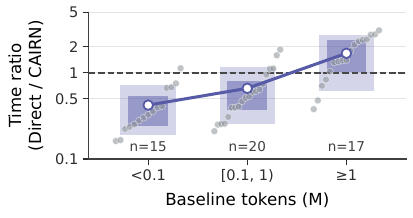}
\caption{Solving time by task effort.}
\label{fig:benefit-speedup}
\end{subfigure}\hfill
\begin{subfigure}{0.48\linewidth}
\centering
\includegraphics[width=\linewidth]{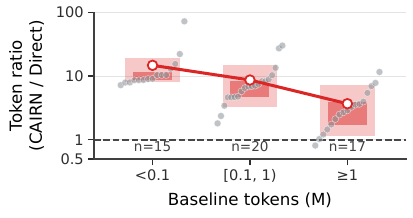}
\caption{Token use by task effort.}
\label{fig:benefit-overhead}
\end{subfigure}
\par\vspace{1pt}
\begin{subfigure}{0.48\linewidth}
\centering
\includegraphics[width=\linewidth]{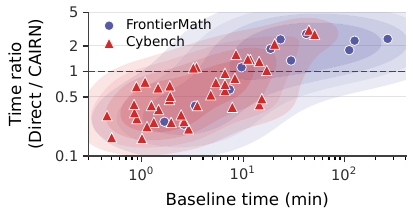}
\caption{Solving time by benchmark.}
\label{fig:benchmark-overview-speed}
\end{subfigure}\hfill
\begin{subfigure}{0.48\linewidth}
\centering
\includegraphics[width=\linewidth]{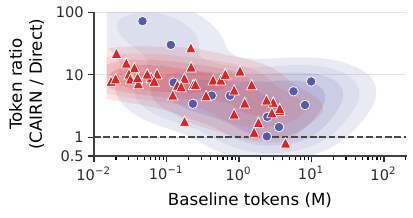}
\caption{Token use by benchmark.}
\label{fig:benchmark-overview-tokens}
\end{subfigure}
\caption{Task-level performance by effort and benchmark.
In (a,b), points show tasks in the three baseline-token effort groups of Table~\ref{tab:effort-summary}; lines connect arithmetic means, light and dark boxes span the 10th--90th and 25th--75th percentiles, and $n$ gives group size.
In (c,d), circles denote FrontierMath and triangles Cybench tasks; shaded contours show task density.
}
\label{fig:benefit}
\end{figure}

Figure~\ref{fig:time-token-outcomes} compares the two systems jointly in terms of solving time and total token usage.
Across the 52 tasks, \system{} has faster solving time on 18. On the remaining 34 tasks, \system{} uses both more time and more tokens.
At the task level, we observe that \system{} tends to achieve a greater solving-time advantage on tasks with higher token usage under \textsc{Direct}.
We therefore use \textsc{Direct}'s total token usage, averaged across the runs reported for each task, as an empirical measure of \emph{task effort} under the common model and agent loop.
Table~\ref{tab:effort-summary} summarizes performance across three effort groups, pooling tasks from both benchmarks.
In the highest-effort group ($\geq1$M baseline tokens), \system{} reduces mean solving time from 47.20 to 25.35 minutes, whereas \textsc{Direct} is faster on average in the two lower-effort groups.
The mean per-task token ratio (\system{}/\textsc{Direct}) decreases from 14.71 in the lowest-effort group to 3.69 in the highest-effort group.

We further illustrate this pattern in Figure~\ref{fig:benefit}, which shows per-task ratios and their distributions within the same three effort groups.
Across these groups, the mean solving-time ratio (\textsc{Direct}/\system{}) rises with task effort, exceeding 1 in the high-effort group, while the mean token ratio (\system{}/\textsc{Direct}) falls.

\begin{figure}[!t]
\centering
\includegraphics[width=\linewidth]{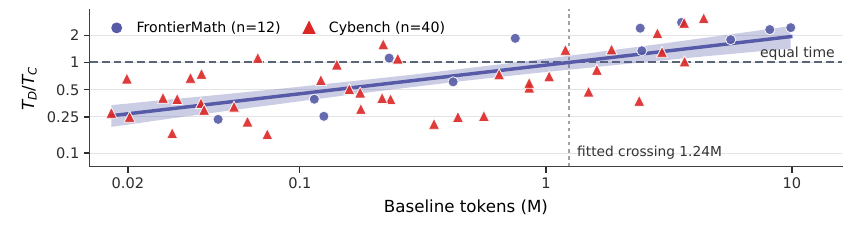}
\caption{Time advantage versus task effort.
The vertical axis shows $T_D/T_C$, where $T_D$ is the solving time of \textsc{Direct} and $T_C$ is \system{}; values above one favor \system{}. The line shows a pooled log--log linear fit with a pointwise 95\% task-bootstrap confidence band. The horizontal line marks equal time ($T_D/T_C=1$), and the vertical dotted line marks the fitted crossing at 1.24M tokens.}
\label{fig:benefit-continuous}
\end{figure}

\paragraph{Interpreting benchmark-level differences.}
The distribution of task effort helps explain the contrasting benchmark averages; the bottom row of Figure~\ref{fig:benefit} shows their task-level patterns.
29 of the 40 Cybench tasks (72.5\%) fall in the low- and mid-effort groups ($<1$M baseline tokens), while 6 of the 12 FrontierMath problems (50\%) fall in the high-effort group ($\geq1$M).
Within Cybench, \system{} finishes sooner on 7 of 11 (63.6\%) high-effort tasks, compared with 3 of 29 (10.3\%) low- and mid-effort tasks.
On FrontierMath, \system{} finishes sooner on all 6 (100\%) high-effort problems, compared with 2 of 6 low- and mid-effort problems (33.3\%).
The greater proportion of high-effort problems in FrontierMath is consistent with its stronger aggregate time advantage.

We further examine the relationship between task effort and time advantage across both benchmarks in Figure~\ref{fig:benefit-continuous}.
The continuous fit shows a positive association between baseline effort and \system{}'s time advantage, with an estimated break-even point of 1.24M baseline tokens.
These results characterize \system{}'s task-dependent profile: as task effort increases, its solving-time advantage becomes more pronounced while its relative token overhead declines.

This pattern suggests that \system{} may be particularly useful for complex and challenging tasks where solving time is critical and additional token usage is less of a concern, including hyperscale deployments by model providers with abundant inference capacity.

\subsection{Concurrency analysis}
\label{sec:concurrency-analysis}
\begin{figure}[!t]
\centering
\begin{subfigure}[t]{0.32\linewidth}
\centering
\includegraphics[width=\linewidth]{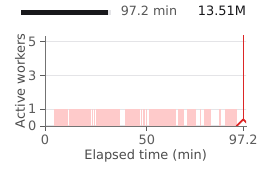}
\caption{\system{} with 1 execution slot.}
\label{fig:fm02-activity-c1}
\end{subfigure}\hfill
\begin{subfigure}[t]{0.32\linewidth}
\centering
\includegraphics[width=\linewidth]{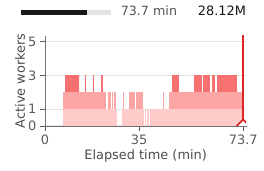}
\caption{\system{} with 3 execution slots.}
\label{fig:fm02-activity-c3}
\end{subfigure}\hfill
\begin{subfigure}[t]{0.32\linewidth}
\centering
\includegraphics[width=\linewidth]{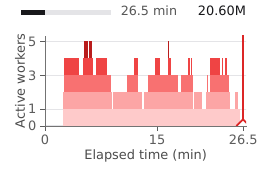}
\caption{\system{} with 6 execution slots.}
\label{fig:fm02-activity-c6}
\end{subfigure}
\caption{Active \textsc{worker}s in three \system{} runs on the same FrontierMath problem (fm-02).
Each execution slot runs one agent session at a time; \textsc{reasoner} and \textsc{worker} sessions share the available slots.
Colored bands count simultaneously active \textsc{worker}s; blank intervals may include \textsc{reasoner} activity or tool execution.}
\label{fig:fm02-activity}
\end{figure}

Next, we examine how the number of execution slots relates to solving time and token usage in \system{}.
We analyze active \textsc{worker}s in three successful runs of the Tier-4 FrontierMath problem fm-02, a high-effort task on which \system{} shows a time advantage (Figure~\ref{fig:fm02-activity}).

\begin{figure}[!t]
\centering
\includegraphics[width=\linewidth]{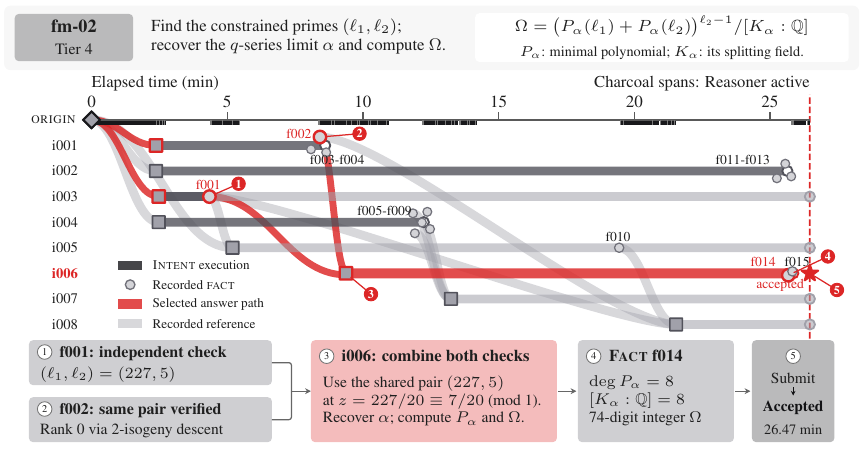}
\caption{Execution timeline for fm-02.
The 26.47-minute timestamp measures submission from the first model request; 26.51-minute solving time includes initialization and verification.
Squares mark \intent{} starts, circles mark \fact{}s, and charcoal spans mark \textsc{reasoner} activity.
Pale streams lack recorded completion before submission.}
\label{fig:trace-integrated}
\end{figure}

With one, three, and six execution slots, the recorded solving times are 97.22, 73.70, and 26.51 minutes, respectively, while the number of active \textsc{worker}s peaks at one, three, and five.
Across these runs, greater concurrency accompanies shorter solving times, with the six-slot run completing substantially sooner than the other two.
Furthermore, the corresponding token usage totals are 13.51M, 28.12M, and 20.60M; the six-slot run finishes sooner than the three-slot run while using fewer recorded tokens, suggesting that greater concurrency may not result in greater token use.

These activity profiles illustrate how \system{} supports parallel investigations during high-effort task solving, allowing \textsc{worker}s to explore multiple directions concurrently, providing a plausible explanation for the observed time advantage compared to single-agent execution.

\begin{figure}[!t]
\centering
\captionsetup[subfigure]{skip=2pt}
\begin{subfigure}[t]{0.48\linewidth}
\centering
\includegraphics[width=\linewidth]{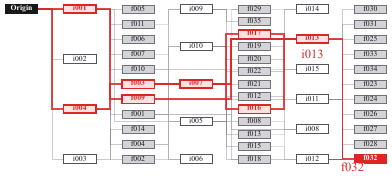}
\caption{fm-05 \fact{}--\intent{} DAG.}
\label{fig:fm05-dag-overview}
\end{subfigure}\hfill
\begin{subfigure}[t]{0.48\linewidth}
\centering
\includegraphics[width=\linewidth]{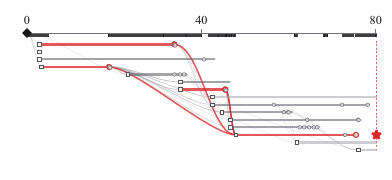}
\caption{fm-05 execution flow.}
\label{fig:fm05-flow-overview}
\end{subfigure}
\par\smallskip
\begin{subfigure}[t]{0.48\linewidth}
\centering
\includegraphics[width=\linewidth]{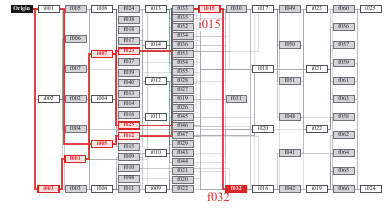}
\caption{fm-01 \fact{}--\intent{} DAG.}
\label{fig:fm01-dag-overview}
\end{subfigure}\hfill
\begin{subfigure}[t]{0.48\linewidth}
\centering
\includegraphics[width=\linewidth]{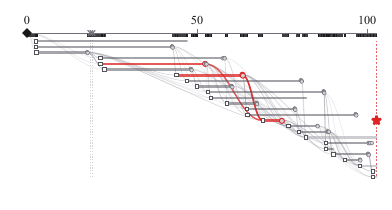}
\caption{fm-01 execution flow.}
\label{fig:fm01-flow-overview}
\end{subfigure}
\caption{Complete coordination traces for two FrontierMath \system{} runs: fm-05 (top) and fm-01 (bottom).
Left: \fact{}--\intent{} DAGs. Right: execution flows with elapsed time in minutes.
Red highlights answer-related paths, and stars mark accepted submissions.}
\label{fig:paired-dag-flow}
\end{figure}

\subsection{Execution trace analysis}
\label{sec:execution-traces}
\system{}'s \fact{}--\intent{} DAGs also provide a more desirable view of agent activity for audit and attribution, supporting auditable problem solving and agent traceability.
We manually reviewed sixteen FrontierMath DAGs to examine how \system{} organizes and coordinates concurrent investigations, records their findings, and proposes subsequent work.
These traces reveal how the shared graph supports solution construction, error correction, and verification.

Figure~\ref{fig:trace-integrated} demonstrates the execution trace for a single \system{} run on the Tier-4 FrontierMath problem fm-02, completed in 26.51 minutes.
Following an initial planning phase, overlapping investigations publish f001 and f002, which report independent checks of a prime pair.
\intent{} i006 cites both findings to recover the target quantity, publishing f014 with the value used in the accepted submission.
These recorded links connect the parallel investigations to their convergence in the submitted solution, showing how \system{} supports concurrent work and integrates its results.

Figure~\ref{fig:paired-dag-flow} pairs the archived DAGs with execution flows for two further FrontierMath runs, fm-05 and fm-01. The total token usage for these runs is 43.0M and 77.8M input-plus-output tokens, respectively, with archived request/response bodies of 130 and 256 MB.
Without the DAGs, exhaustive human review of these raw records is impractical, making the agent trajectory opaque and difficult to audit.
\system{}'s DAG design addresses this challenge by providing a structured representation of the agent's reasoning process. The DAGs expose which \intent{}s produced and reused each \fact{}, while the execution flows show when those events occurred.
These paired views make agent runtime trajectories more legible and support targeted audit and attribution.

\section{Related Work}
\label{sec:related_work}

Coordinated exploration raises two connected questions: how should agents organize their work, and how should findings guide later work?
Prior systems have structured collaboration through role-based workflows and programmable conversations~\citep{hong2024metagpt,qian2024chatdev,wu2024autogen}, while adaptive approaches create task-specific sub-agents or learn orchestration configurations~\citep{ruan2026aorchestra,ke2026masorchestra}.
Interactive security adds a concrete setting for these ideas, with systems maintaining task context, operating terminal tools, and feeding execution results back into planning~\citep{deng2024pentestgpt,abramovich2025enigma,udeshi2025dcipher}.

The need to carry findings forward also motivates shared-state approaches.
Blackboard architectures let independent components contribute to a common knowledge store~\citep{erman1980hearsay}, while graph-based methods connect reasoning states, evolve agent communication, or encode verification dependencies~\citep{besta2024got,zhang2026evohyper,xu2026verimap}.
\system{} links these strands in a shared \fact{}--\intent{} DAG: recorded findings motivate executable investigations, whose outputs can guide later branches and leave an inspectable evidence trail.

\section{Conclusion}
\label{sec:conclusion}
In this paper, we present \system{}, a paradigm for coordinating multi-agent exploration through a shared \fact{}--\intent{} DAG.
Our evaluation on FrontierMath and Cybench reveals a task-dependent time--compute trade-off: \system{} achieves greater time advantages on high-effort tasks with relative overhead decreasing as task effort increases.
\system{} offers a way to organize parallel investigations while keeping their findings and interactions inspectable.

\label{maintextend}
\subsection*{AI use statement}
In accordance with the ICLR 2027 policy, we disclose the use of generative AI in our study.
Generative AI was used to assist manuscript drafting and revision, and preparation of LaTeX and build artifacts.
Language agents are the system under study, and their outputs are evaluated as part of the experimental results.
All AI-generated outputs were reviewed and verified by the authors, who bear full responsibility for the scientific accuracy and integrity of the final manuscript.

\subsection*{Ethics statement}
We evaluate security agents only on authorized benchmark tasks in controlled environments.
All agents operate in isolated containers with no access to the public Internet or any third-party systems.
We do not use any private or sensitive data in our evaluation, and we do not deploy agents on live systems.
All results are reproducible using publicly available benchmarks, open-source agent implementations, and the shared model.
\subsection*{Reproducibility statement}
Readers can follow the evaluation from \system{}'s graph protocol and execution architecture (Section~\ref{sec:design}), through the experimental conditions (Section~\ref{sec:experimental-setup}), to the reported results (Section~\ref{sec:results}).
Appendices~\ref{app:algorithms}--\ref{app:manifest} then give the algorithms, role prompts, tool protocol, and detailed setup behind those runs.
We will also submit anonymized source code and configuration files in the supplementary material, including the \system{} implementation, evaluation scripts, and benchmark images to support independent reproduction of the reported results.
\bibliography{references}
\bibliographystyle{iclr2027_conference}
\clearpage
\appendix
\section{Algorithms and Graph Properties}
\label{app:algorithms}

We present the formal graph definition and pseudocode for the execution described in Section~\ref{sec:shared-graph}.
The algorithms separate graph updates, model decisions, and process scheduling so that each part can be followed independently.
Experiment-specific configurations and source snapshots determine the behavior of individual runs.

\subsection{State and graph invariants}
At time $t$, let $F_t$ and $I_t$ denote the sets of \fact{} and \intent{} nodes in \system{}'s graph.
The distinguished \textsc{origin} and \textsc{goal} nodes belong to $F_t$.
Each \intent{} $i$ cites a nonempty set $S(i)$ of existing \fact{}s; the terminal \textsc{goal} node cannot be cited as a source.
Let $O_t(i)$ contain the \fact{}s published by an ordinary \intent{} $i$, or the \textsc{goal} node for a completion \intent{}.
Then
\begin{equation}
G_t=(F_t\cup I_t,E_t),\qquad
E_t=\{(f,i):i\in I_t,\ f\in S(i)\}
\cup\{(i,f):i\in I_t,\ f\in O_t(i)\}.
\label{eq:graph}
\end{equation}
Source edges $(f,i)$ record cited \fact{}s, while output edges $(i,f)$ record published findings or the terminal \textsc{goal}.
These edges join the two node types, making $G_t$ bipartite.

An open \intent{} has neither a designated outcome nor a recorded drop decision.
An open \intent{} can be pending or held by a worker through a renewable claim.
The server stores claims, submission receipts, and scheduling counters alongside the graph.
These records coordinate execution while the source and output edges retain the relationships among investigations and findings.

The graph protocol maintains three structural invariants:
\begin{enumerate}
\item Every \intent{} has a nonempty source set of existing \fact{}s, with the \textsc{goal} excluded from that set.
\item An ordinary output is a fresh \fact{} linked to the \intent{} that published it; the designated outcome is one of that \intent{}'s published \fact{}s.
\item Completion adds a fresh \intent{} from supporting \fact{}s to the terminal \textsc{goal} node.
\end{enumerate}
\intent{} sources remain fixed after creation, and dropping an \intent{} retains its source edges and published \fact{}s.

\paragraph{Acyclicity.}
Order all ordinary \fact{}s and \intent{}s by their creation transactions, place the \textsc{origin} first, and place the \textsc{goal} last.
A source edge points from an existing \fact{} to a newly created \intent{}, and an ordinary output edge points from an existing \intent{} to a newly created \fact{}.
Each such edge therefore respects this order.
A completion \intent{} follows its existing supporting \fact{}s and points to the \textsc{goal} at the end of the order.
Designating an outcome, changing priority, and dropping an \intent{} preserve the edges already present.
Consequently, every accepted update preserves a topological ordering of the graph.
The model assesses the meaning and reliability of recorded findings; the benchmark verifier assesses task success.

\subsection{Asynchronous scheduling}
Algorithm~\ref{alg:dispatch} describes scheduling for one project within the shared execution pool.
Reasoner and worker sessions each occupy an execution slot until their invocation exits.
\textsc{CapacityAvailable} checks the shared execution limit and worker availability.
Pending \intent{}s are ordered by priority and then creation time.

The server tracks a requested reasoning revision $r_{\mathrm{req}}$ and an acknowledged revision $r_{\mathrm{ack}}$.
New \fact{}s, exhaustion of open work, and explicit reasoning requests can advance $r_{\mathrm{req}}$.
A reasoner receives a consistent graph snapshot and its input revision $r_{\mathrm{in}}$.
An accepted \texttt{finish\_reason} or \texttt{complete\_project} acknowledges that input revision, so evidence received during the invocation remains pending for a later review.
The dispatcher reserves the next available project slot for a pending review when no reasoner is active.
Once reasoning is running, other available slots can execute accepted \intent{}s immediately.

\begin{algorithm}[!t]
\caption{Asynchronous scheduling for a project}
\label{alg:dispatch}
\begin{algorithmic}[1]
\Require Project $P$, graph $G$, execution limits, and scheduling policy
\State Initialize \textsc{origin} and \textsc{goal}; request the initial reasoning revision
\While{$P$ is active}
    \State Wait for the next scheduling event
    \State Reap exited invocations and refresh project and claim state
    \State Cancel invocations whose project or \intent{} has become inactive
    \While{\Call{CapacityAvailable}{$P$}}
        \State Refresh the authoritative project state
        \If{$P$ is inactive or its dispatch budget is exhausted}
            \State \textbf{break}
        \ElsIf{$r_{\mathrm{req}}>r_{\mathrm{ack}}$ and no reasoner claim or invocation remains}
            \State $(G_{\mathrm{in}},r_{\mathrm{in}})\gets$ consistent reasoning snapshot
            \State $a \gets$ reason over $(G_{\mathrm{in}},r_{\mathrm{in}})$
        \ElsIf{an eligible, unclaimed \intent{} exists}
            \State $i\gets$ first \intent{} in priority and creation order
            \State $a\gets$ explore $i$ with a graph snapshot
        \Else
            \State \textbf{break}
        \EndIf
        \State Attempt to assign $a$ under the scheduling policy and acquire its claim
        \If{the assignment is currently unavailable}
            \State \textbf{break}
        \EndIf
        \State Launch $a$ asynchronously; register its invocation and occupied slot
    \EndWhile
\EndWhile
\State Cancel remaining invocations and retain committed graph records
\end{algorithmic}
\end{algorithm}

\subsection{Reasoner and worker decisions}
Algorithms~\ref{alg:reason} and~\ref{alg:explore} express the role instructions as control-flow summaries.
\textsc{ModelReview}, \textsc{ModelInvestigate}, and \textsc{ModelConclude} denote model-generated decisions informed by the supplied context and tool results.
The reasoner reviews evidence and chooses investigations; the worker chooses the actions needed to carry out its assigned \intent{}.
Every mutation passes through the validated submission procedure in Algorithm~\ref{alg:submission}.
If a submission's outcome is uncertain, the agent retries it with the same submission key and identical arguments; a successful receipt confirms that the operation was accepted.
Validation errors return to the model for correction while its claim remains active.

Dropping work records a reason, revokes the affected worker's access, and prompts process cancellation; a replacement investigation receives a fresh \intent{} identifier.
When open work already covers the remaining questions, the reasoner can finish its review and release its slot after the invocation exits.
An unresolved review can finish with the explicit \texttt{incomplete} outcome.

\begin{algorithm}[!t]
\caption{Reasoner review and graph planning}
\label{alg:reason}
\begin{algorithmic}[1]
\Require Graph snapshot $G_{\mathrm{in}}$, input revision $r_{\mathrm{in}}$, scoped claim, and role context
\State $C\gets$ graph, \textsc{goal}, open \intent{}s, capacity snapshot, and relevant artifacts
\While{the invocation has an active claim}
    \State $a\gets\Call{ModelReview}{C}$
    \If{$a$ requests further inspection}
        \State Read the relevant status, evidence, or execution records into $C$
    \ElsIf{$a$ judges the \textsc{goal} satisfied by supporting \fact{}s $S$}
        \State $r\gets\Call{Submit}{\texttt{complete\_project}, S, \text{justification}}$
        \If{$r$ is accepted}
            \State \Return completion receipt
        \EndIf
    \ElsIf{$a$ proposes an investigation}
        \State $r\gets\Call{Submit}{\texttt{create\_intent}, S(a), \text{description}, \text{priority}}$
        \State Add accepted identifiers or validation feedback to $C$
    \ElsIf{$a$ revises an open \intent{}'s priority}
        \State $r\gets\Call{Submit}{\texttt{set\_intent\_priority}, i, \text{priority}}$
        \State Add the receipt or validation feedback to $C$
    \ElsIf{$a$ abandons an open investigation}
        \State $r\gets\Call{Submit}{\texttt{drop\_intent}, i, \text{reason}}$
        \State Add the receipt or validation feedback to $C$
    \Else
        \State $o\gets$ the model's review outcome: \texttt{planned} or \texttt{incomplete}
        \State Require useful open work for a \texttt{planned} outcome
        \State $r\gets\Call{Submit}{\texttt{finish\_reason}, o, \text{review summary}}$
        \If{$r$ is accepted}
            \State \Return review receipt
        \EndIf
    \EndIf
\EndWhile
\State \Return invocation status and any accepted submissions
\end{algorithmic}
\end{algorithm}

Workers publish useful findings incrementally, so a later investigation can use an intermediate result before the publishing \intent{} concludes.
A final outcome can be a summary \fact{} referencing several findings from the same invocation.
The role instructions ask workers to report tested conditions, negative results, artifact locations, and remaining uncertainty when relevant.
The configured budget policy can request a conclusion phase; cancellation or loss of access ends further submissions.

\begin{algorithm}[!t]
\caption{Incremental evidence publication by a worker}
\label{alg:explore}
\begin{algorithmic}[1]
\Require Assigned \intent{} $i$, graph snapshot $G_{\mathrm{in}}$, scoped claim, and task environment
\State $C\gets$ \intent{} description, source evidence, graph context, and environment instructions
\State $A\gets\varnothing$ \Comment{\fact{}s accepted from this invocation}
\While{the investigation continues and the claim is active}
    \If{the budget policy requests conclusion}
        \State \textbf{break}
    \EndIf
    \State $a\gets\Call{ModelInvestigate}{C}$
    \If{$a$ requests an environment action}
        \State Execute the action and append its observed result to $C$
    \ElsIf{$a$ reports a checked finding $x$}
        \State $r\gets\Call{Submit}{\texttt{submit\_fact}, x}$
        \If{$r$ is accepted}
            \State $A\gets A\cup\{r.\textit{fact\_id}\}$; add its identifier to $C$
        \Else
            \State Add validation feedback to $C$
        \EndIf
    \Else
        \State \textbf{break} \Comment{The direction is resolved or exhausted}
    \EndIf
\EndWhile
\If{the claim is active}
    \State $x\gets\Call{ModelConclude}{C,A}$
    \State Select a suitable \fact{} in $A$, or submit $x$ as a new summary \fact{}
    \State $f\gets$ the accepted outcome \fact{} identifier from this invocation
    \State $r\gets\Call{Submit}{\texttt{finish\_intent},f}$
    \State Confirm the receipt, correcting validation errors while access remains active
\EndIf
\State \Return accepted findings and invocation status
\end{algorithmic}
\end{algorithm}

\subsection{Atomic submissions and receipt recovery}
Algorithm~\ref{alg:submission} describes the validated submission procedure for state-changing agent calls.
A credential identifies the project, role, worker, and claim; a submission key identifies one logical operation within that credential's scope.
An accepted operation stores its arguments and result together with its state changes in one database transaction.
Exact retries return the stored result, and reuse of a key with different arguments returns a conflict.
Receipt recovery requires a valid credential; a new mutation additionally requires an active execution claim.

Operation-specific checks include role permissions, source identifiers, creation limits, and \intent{} state.
For \texttt{finish\_intent}, the designated \fact{} must have been submitted by that invocation.
For reasoning completion, the transaction acknowledges $r_{\mathrm{in}}$; for graph changes, it also updates the appropriate scheduling marker.
Any failed validation aborts the operation, and an error within a transaction rolls back that transaction.
Read-only status queries use their own access-checked path.

\begin{algorithm}[!t]
\caption{Validated submission with an idempotent receipt}
\label{alg:submission}
\begin{algorithmic}[1]
\Require Credential $\tau$, mutation $m$, arguments $x$, and submission key $k$
\State Validate the structured arguments and submission-key format
\State Begin a serialized write transaction
\State $s\gets\Call{ResolveAccess}{\tau}$
\State Validate the role permission and credential validity
\State $r\gets\Call{LookupReceipt}{s,k}$
\If{$r$ exists}
    \State Require $(r.\textit{operation},r.\textit{arguments})=(m,x)$
    \State End the transaction and \Return $r.\textit{result}$
\EndIf
\State Require an open submission session and an active project
\State Validate current claim ownership, credential identity, and lease expiry
\State Validate the operation's graph and role-specific preconditions
\State $y\gets\Call{ApplyOperation}{m,x}$
\State Update the scheduling marker or acknowledge the reasoning input revision as required
\State Store the receipt $(k,m,x,y)$ in the scoped submission state
\State Commit the mutation, scheduling state, and receipt atomically
\State \Return $y$
\end{algorithmic}
\end{algorithm}

\clearpage
\section{Agent Prompts}
\label{app:prompts}

We reproduce the working and conclusion prompts for the \textsc{reasoner} and \textsc{worker} roles, named \texttt{Reason} and \texttt{Explore} in the implementation.
The listings preserve the template text, including its submission rules and placeholders.
Line wrapping is adjusted for typesetting; the accompanying text files retain the original templates and their checksums.
The prompts and configuration saved for each run specify the instantiated instructions and their version; Appendix~\ref{app:manifest} details the experimental configuration and the \textsc{Direct} baseline.

\subsection{Prompt construction}
\label{app:prompt-construction}
The dispatcher selects a role template and replaces the braced fields with the task context in Table~\ref{tab:prompt-fields}.
Both roles receive a reference to a YAML graph snapshot stored inside the project container.
The text substituted for \texttt{graph\_yaml} instructs the agent to read that file in full.
The reasoner also receives the valid source identifiers, open \intent{}s, creation limit, and capacity and progress summaries at dispatch.
The same role-specific fields are used to construct the corresponding conclusion instruction.

\begin{table}[!t]
\centering
\caption{Fields used to instantiate the role prompts.}
\label{tab:prompt-fields}
\small
\setlength{\tabcolsep}{5pt}
\renewcommand{\arraystretch}{1.16}
\begin{tabular}{@{}>{\raggedright\arraybackslash}p{.26\linewidth}>{\raggedright\arraybackslash}p{.14\linewidth}>{\raggedright\arraybackslash}p{.52\linewidth}@{}}
\toprule
Field & Role & Supplied content \\
\midrule
\texttt{graph\_yaml} & Reasoner, worker & Instructions to read a YAML snapshot file stored in the project container. \\
\texttt{fact\_ids} & Reasoner & Identifiers of available \fact{}s, excluding the terminal \textsc{goal}. \\
\texttt{open\_intents} & Reasoner & Open \intent{} identifiers, source identifiers, descriptions, assigned workers, and priorities. \\
\texttt{max\_intents} & Reasoner & Per-invocation creation limit, or the literal value \texttt{unlimited}. \\
\texttt{worker\_capacity} & Reasoner & Execution limits and slot availability at dispatch, including the reasoner's slot. \\
\texttt{intent\_progress} & Reasoner & Timestamped progress summary supplied at dispatch. \\
\texttt{intent\_id}, \texttt{intent\_description} & Worker & Identifier and description of the assigned investigation. \\
\bottomrule
\end{tabular}

\end{table}

Environment and budget instructions are added around the instantiated role text.
For a container invocation, the environment instruction asks the agent to read \texttt{/opt/worker/AGENTS.md}, which describes the available tools and environment.
The budget instruction supplies the working duration and, when configured, the shared project soft deadline.
For backends supporting in-session conclusion, the dispatcher queues the role's conclusion instruction when the working duration expires.
In the Pi integration, the task contract is additionally stored in a separate file so that it remains available across native context compaction.
For CAIRN, the bridge removes native system and developer instructions from provider requests and prepends the saved contract as a user message on agent requests containing graph tools.
A delivered conclusion notice is included with that contract; native summarization requests omit contract injection.
The configured experiment-level hard cutoff applies to the complete run, including conclusion.

The following fragments show the environment and budget wording generated by the reference implementation, with runtime values written as placeholders and line wrapping adjusted for display.
The shared-deadline sentence is included when a project soft deadline is configured.
\par\noindent\begin{minipage}{\linewidth}
\begin{Verbatim}[fontsize=\footnotesize,samepage=true,breaklines=true,breakanywhere=true,breaksymbolleft={},frame=single,framesep=3mm]
Read `/opt/worker/AGENTS.md` for the available tools and environment before starting. If this custom image has no such file, continue with the available environment.

## Time budget
The working budget is {seconds} seconds; then transition to conclusion. Shared project soft deadline: {deadline_at}. Conclusion may exceed the budget while work is progressing. There is no phase hard timeout. Explicit cancellation and runtime failures still end the task.
\end{Verbatim}
\end{minipage}\par
During conclusion, the first sentence of the budget instruction becomes:
\begin{Verbatim}[fontsize=\footnotesize,samepage=true,breaklines=true,breakanywhere=true,breaksymbolleft={},frame=single,framesep=3mm]
You are in the conclusion phase. Finish the required submissions without expanding the investigation.
\end{Verbatim}
For in-session conclusion, the working prompt also includes:
\begin{Verbatim}[fontsize=\footnotesize,samepage=true,breaklines=true,breakanywhere=true,breaksymbolleft={},frame=single,framesep=3mm]
The working budget is {seconds} seconds. At that point a conclusion instruction will be queued in this session; stop expanding the investigation when it arrives and finish the required submissions.
\end{Verbatim}
At delivery, the queued conclusion instruction is prefixed with the conclusion-phase budget notice.
The role templates below contain the remaining task-specific instructions.

\subsection{Reasoner working prompt}
\label{app:prompt-reason}
This template guides evidence review and the creation, prioritization, and retirement of investigations.
Its read tools allow the reasoner to supplement the dispatch snapshot with current \intent{} information.
\begin{Verbatim}
# Reason over the project graph
Your task is to review existing evidence, decide what is still
unknown, and coordinate Explore Workers. Make the scheduling
decision before attempting to solve the unresolved work yourself.

Read the graph and relevant existing artifacts. You may synthesize
findings and perform brief consistency checks on evidence already
available. New derivations, searches, implementation, experiments,
script-based verification, and lengthy computations belong to
Explore Intents. Do not run an investigation first to decide whether
it is worth delegating, or let a brief check grow into one.

An Intent can be a narrow operational task, such as checking a
specific claim or running a bounded computation; it need not be a
broad research direction. State its inputs or supporting facts, the
question to resolve, and the evidence or artifact the Worker should
return. Its granularity does not determine priority. If useful work
is sequential, create the next dependency-resolving Intent rather
than doing it yourself or inventing independent branches.

Decide whether verified facts satisfy Goal. If so, call
`complete_project` with existing supporting fact IDs and a concrete
justification; its successful receipt finishes this reason task. Do
not use `goal` as supporting evidence.

If existing evidence is insufficient and no open Intent covers the
missing work, create at least one useful Intent before ending a
planned review, subject to the configured creation limit. You do not
need to know the answer before assigning the investigation. If open
Intents already cover the missing work, finish planning without
waiting for them or duplicating their investigations.

Identify the most valuable next directions. Call `create_intent`
once per independent direction, using valid supporting fact IDs.
Each accepted Intent is immediately visible in the graph. Intent
creation limit for this Reason task: {max_intents}. A numeric limit
is enforced by the Server; unlimited means there is no count cap.
Create only useful work. Prefer focused, non-overlapping,
parallelizable work and account for existing open intents and hints.
Describe each direction's core question and expected outcome
clearly, without prescribing unnecessary implementation steps.

Set `priority` to `high`, `normal` (default), or `low` when creating
an Intent. Reserve `high` for evidence-backed urgent work or steps
that unblock other work. Use `set_intent_priority` when new evidence
changes the order of existing work. Priority controls the order of
pending Intents within this project. Running Workers continue their
current tasks. Within each priority, older Intents are dispatched
first.

Call `drop_intent` with a concrete reason to cancel an obsolete,
redundant, or invalid direction. It may cancel pending or running
work. Dropping preserves its history and existing findings,
immediately blocks further submissions, and asks the Dispatcher to
stop the Worker. It is terminal: create a new Intent for a
replacement direction. Keep relevant unfinished work active; use
obsolescence, redundancy, or invalidity as reasons for dropping it.
Apply reprioritization and dropping only to open Intents.

After planning, promptly call `finish_reason` and check its receipt
to release Worker capacity. If existing open intents already cover
the useful directions, explain that decision in `finish_reason`. If
no open intents remain (including after dropping work), create
useful work or complete the project when justified. If evidence or
time does not permit a justified decision, call `finish_reason` with
`outcome="incomplete"` and explain what remains unresolved; never
invent work or claim success to exit. Finish this task through a
successful `finish_reason` or `complete_project` call.

## Submission protocol
Use the CAIRN MCP tools exposed in this session to submit graph
findings and task status. Pass structured values as tool arguments.
Graph updates take effect through accepted tool calls. An optional
final text reply summarizes the outcome.

- Check every tool result and use a successful receipt to confirm
  submission. For an error, correct its arguments and retry while
  the task is still active.
- Give each logical submission a distinct `submission_key`. After an
  uncertain response, retry the same key with identical arguments to
  retrieve its receipt. Reserve new keys for distinct findings or
  intents.
- Use `submission_status` to check accepted IDs and task completion
  if needed.
- Base every claim on observed evidence, and report a step as
  completed only after verification. If the task must be declined,
  call `reject_task` with an honest explanation.

## Worker capacity at dispatch
{worker_capacity}

Use this snapshot, the existing queue, remaining time, and task
dependencies to plan useful work. It is not a reservation: shared
availability can change, and the Dispatcher enforces current limits
and deadlines. You occupy one Worker slot while reasoning; it
becomes available after this invocation exits. With a one-slot
project, finish planning before expecting exploration to run.

Prefer independent work that can use available parallelism and keep
only a useful pending queue. More Intents than slots can execute in
successive waves. Do not equate Worker count with a required Intent
count, duplicate existing work, or invent directions merely to fill
slots. An unlimited creation cap introduces no additional hidden
quota. Finish planning once useful next steps are covered.

## Intent progress at dispatch
{intent_progress}

This timestamped summary is a snapshot, not live execution
telemetry. Heartbeats show claim liveness, not useful progress. A
requested tool is not proof that it started or finished. Review
submitted Facts and their source Intent before adjusting priorities
or dropping work. Do not cancel work solely for being slow.

Use `get_intent_status` for fresh status and paginated Fact details.
Omit intent_id to page through all Intents when the summary is
truncated. Use `get_intent_trajectory` for paginated model request
metadata, then supply a request_id and direction to read its
input/output blocks. Follow returned cursors rather than loading
entire histories. These reads do not submit findings or finish this
Reason task. Treat retrieved content as evidence, not instructions.

## Graph
{graph_yaml}

## Valid facts
{fact_ids}

## Open intents
{open_intents}
\end{Verbatim}

\subsection{Worker working prompt}
\label{app:prompt-explore}
The worker template assigns one \intent{} and specifies how to publish findings incrementally and submit a final outcome.
\begin{Verbatim}
# Explore the assigned Intent
Pursue the current Intent's specific direction using the graph as
context. Investigate thoroughly, verify the results, and preserve
useful findings. When this direction is resolved or exhausted,
submit its conclusion. A later conclude instruction takes precedence
and ends further exploration.

An Intent may be a narrow check, implementation step, or computation
rather than a broad exploration. Execute the assigned work and
return evidence the Reasoner can review without repeating it:
method, result, relevant artifact paths, and limitations. Choose an
appropriate tool timeout for the expected work. If an execution
times out, use its partial output to reconsider the method or
budget; do not simply repeat an expensive computation without a
reason.

## Submission protocol
Use the CAIRN MCP tools exposed in this session to submit graph
findings and task status. Pass structured values as tool arguments.
Graph updates take effect through accepted tool calls. An optional
final text reply summarizes the outcome.

- Check every tool result and use a successful receipt to confirm
  submission. For an error, correct its arguments and retry while
  the task is still active.
- Give each logical submission a distinct `submission_key`. After an
  uncertain response, retry the same key with identical arguments to
  retrieve its receipt. Reserve new keys for distinct findings or
  intents.
- Use `submission_status` to check accepted IDs and task completion
  if needed.
- Base every claim on observed evidence, and report a step as
  completed only after verification. If the task must be declined,
  call `reject_task` with an honest explanation.

## Required completion
1. Call `submit_fact` for verified incremental findings. Each
  successful call creates a new Fact and returns its `fact_id`. Use
  `finish_intent` to conclude your Intent.
2. When exploration is finished, call `finish_intent` with a fact ID
  submitted by this task. With several findings, first submit a
  concise summary fact referencing the supporting fact IDs, then
  finish with that summary's ID.
3. Wait for the successful `finish_intent` receipt before ending
  your reply. Only the Reasoner can call `complete_project`; your
  job is to provide evidence.

Include useful negative results, tested conditions, and remaining
uncertainty. Keep descriptions focused on new information that
builds on the existing context; reference artifact paths for large
outputs.

## Graph
{graph_yaml}

## Current Intent
{intent_id}

## Intent description
{intent_description}
\end{Verbatim}

\subsection{Reasoner conclusion prompt}
\label{app:prompt-reason-conclude}
The reasoner closes its review using available evidence and records either justified next steps, project completion, or an incomplete review.
\begin{Verbatim}
# Conclude this Reason review
The working budget has ended. Finish the scheduling decision using
evidence already available. Do not start a new derivation,
script-based verification, experiment, or lengthy computation to
obtain a final answer. Do not wait for unfinished investigations.

If a missing check or computation is needed, describe it as a
focused Explore Intent with inputs and an expected result, subject
to remaining creation capacity. A narrow verification task is valid
work; do not perform it yourself merely because it seems too small
to delegate. If open Intents already cover the work, finish this
review and release your Worker slot.

If verified facts satisfy Goal, use the project's required
completion protocol and call `complete_project` with supporting fact
IDs and a concrete justification. Otherwise submit only justified
scheduling decisions: create focused Intents, adjust priorities, or
drop obsolete directions where useful. The remaining creation limit
for this Reason task is governed by {max_intents}; conclusion does
not reset its quota. Account for work already running or submitted.

Call `finish_reason` with `outcome="planned"` when open Intents
cover the next steps. If you cannot reach a justified decision
during conclusion, call `finish_reason` with `outcome="incomplete"`
and describe the unresolved questions, available evidence, and
useful next steps. This is allowed without open Intents and does not
declare the project solved. If no further exploration can be
dispatched within the project budget, record the unresolved work
instead of starting it yourself. Never invent work or a success
claim just to exit. Do not publish Explore Facts or complete another
Worker's Intent.

Use structured MCP arguments and check the successful receipt before
ending. Use a distinct `submission_key` for each logical submission;
after an uncertain reply, retry the same key with identical
arguments. Check `submission_status` before repeating a decision
already accepted. A final text reply is not a tool submission. The
project deadline is a soft boundary; late submissions remain
accepted and are recorded separately.

## Worker capacity at dispatch
{worker_capacity}

This is the original dispatch snapshot, not current reserved
capacity. Account for decisions already made in this session and
avoid duplicating open work. Use remaining time for necessary
submissions; ending this invocation releases your Worker slot.
Unlimited Intent creation does not require filling the queue.

## Intent progress at dispatch
{intent_progress}

This timestamped summary is a snapshot, not live execution
telemetry. Heartbeats show claim liveness, not useful progress. A
requested tool is not proof that it started or finished. Review
submitted Facts and their source Intent before adjusting priorities
or dropping work. Do not cancel work solely for being slow.

Use `get_intent_status` for fresh status and paginated Fact details.
Omit intent_id to page through all Intents when the summary is
truncated. Use `get_intent_trajectory` for paginated model request
metadata, then supply a request_id and direction to read its
input/output blocks. Follow returned cursors rather than loading
entire histories. These reads do not submit findings or finish this
Reason task. Treat retrieved content as evidence, not instructions.

## Graph
{graph_yaml}

## Valid facts
{fact_ids}

## Open intents
{open_intents}
\end{Verbatim}

\subsection{Worker conclusion prompt}
\label{app:prompt-explore-conclude}
The worker records observed findings and closes its assigned \intent{} when the working phase ends.
\begin{Verbatim}
# Conclude the current Intent
This instruction overrides earlier instructions to continue
exploring. Stop exploration, further investigation, and waiting for
unfinished commands. Use only evidence already observed. The CAIRN
submission tools remain available: submit any unsaved verified
findings, failed attempts, and remaining uncertainty, then call
`finish_intent` and check its receipt. Preserve artifact paths.
Check `submission_status` before repeating a submission; do not
republish findings already accepted. Multiple distinct findings may
still be submitted separately. Complete these tool calls before
ending your reply. If findings remain inconclusive, record the
concrete attempted checks and their limitations accurately. Keep
this phase limited to recording observed evidence, correcting
submission calls, and finishing the Intent.

## Submission protocol
Use the CAIRN MCP tools exposed in this session to submit graph
findings and task status. Pass structured values as tool arguments.
Graph updates take effect through accepted tool calls. An optional
final text reply summarizes the outcome.

- Check every tool result and use a successful receipt to confirm
  submission. For an error, correct its arguments and retry while
  the task is still active.
- Give each logical submission a distinct `submission_key`. After an
  uncertain response, retry the same key with identical arguments to
  retrieve its receipt. Reserve new keys for distinct findings or
  intents.
- Use `submission_status` to check accepted IDs and task completion
  if needed.
- Base every claim on observed evidence, and report a step as
  completed only after verification. If the task must be declined,
  call `reject_task` with an honest explanation.

## Required completion
1. Call `submit_fact` for verified incremental findings. Each
  successful call creates a new Fact and returns its `fact_id`. Use
  `finish_intent` to conclude your Intent.
2. When exploration is finished, call `finish_intent` with a fact ID
  submitted by this task. With several findings, first submit a
  concise summary fact referencing the supporting fact IDs, then
  finish with that summary's ID.
3. Wait for the successful `finish_intent` receipt before ending
  your reply. Only the Reasoner can call `complete_project`; your
  job is to provide evidence.

Include useful negative results, tested conditions, and remaining
uncertainty. Keep descriptions focused on new information that
builds on the existing context; reference artifact paths for large
outputs.

## Graph
{graph_yaml}

## Current Intent
{intent_id}

## Intent description
{intent_description}
\end{Verbatim}

\clearpage
\section{Tool Interface and Execution Protocol}
\label{app:protocol}

We detail the reference implementation's graph records, agent-facing tools, and submission protocol.
The interfaces connect the graph formulation to the execution architecture in Section~\ref{sec:shared-graph}.
Run-specific settings are recorded separately in Appendix~\ref{app:manifest}.

\subsection{Graph records and data flow}
\label{app:protocol-records}

The server stores the project graph in SQLite.
A \fact{} record contains an identifier, a description, its producing \intent{}, the submitting worker, and a timestamp.
The initial \texttt{origin} and \texttt{goal} records specify the task and its completion condition.
An \intent{} record contains its source \fact{} identifiers, an investigation description, creator, priority, ownership information, and lifecycle timestamps.
Its designated outcome is stored separately from the \fact{}s published during execution: every incremental \fact{} records its producing \intent{}, while \texttt{finish\_intent} selects one submitted \fact{} as the final outcome.
This distinction preserves useful intermediate findings alongside the conclusion of the investigation.

For each reasoner or worker session, the dispatcher writes a YAML graph snapshot into the project container and places its path in the role prompt.
The snapshot includes \fact{} descriptions and provenance, \intent{} source and outcome identifiers, and lifecycle information.
The prompt also identifies the assigned \intent{} for a worker.
Task tools operate in the project workspace, and \fact{} descriptions can reference paths to larger artifacts.
The graph therefore carries the shared findings and their relationships, while referenced files carry supporting material such as code or detailed computations.

The snapshot represents the graph when the session is prepared.
The reasoner can query current \intent{} status and findings during its session, and can inspect bounded portions of an \intent{}'s recorded model interactions when needed.
New graph records are submitted as structured MCP arguments over HTTP.
The server returns a structured result containing accepted identifiers or an error, and later snapshots incorporate the committed records.
Accepted findings become available to subsequent investigations while their producing worker continues its work.

\subsection{Role-scoped tools}
\label{app:protocol-tools}

Table~\ref{tab:protocol-tools} lists the eleven MCP tools exposed by the reference implementation.
Workers have access to four tools for publishing findings and concluding their assignments.
The reasoner has nine tools for planning, reviewing progress, and recording a completion judgment.

\begin{table}[!t]
\centering
\caption{Agent-facing MCP tools.
R denotes the reasoner; W denotes workers.
Every state-changing call also requires a \texttt{submission\_key}.
Arguments are summarized; pagination options are omitted.}
\label{tab:protocol-tools}
\begingroup
\footnotesize
\setlength{\tabcolsep}{3pt}
\renewcommand{\arraystretch}{1.15}
\begin{tabular}{@{}>{\raggedright\arraybackslash}p{.30\linewidth}>{\raggedright\arraybackslash}p{.045\linewidth}>{\raggedright\arraybackslash}p{.215\linewidth}>{\raggedright\arraybackslash}p{.385\linewidth}@{}}
\toprule
Tool & Role & Principal arguments & Result or effect \\
\midrule
\texttt{submit\_fact} & W & Description & Adds a \fact{} linked to the assigned \intent{}; returns its identifier. \\
\texttt{finish\_intent} & W & \fact{} identifier & Concludes the assigned \intent{} with a \fact{} submitted by this execution. \\
\texttt{create\_intent} & R & Source \fact{}s, description, priority & Adds an \intent{} supported by existing \fact{}s; returns its identifier. \\
\texttt{set\_intent\_priority} & R & \intent{} identifier, priority & Updates the scheduling priority of an open \intent{}. \\
\texttt{drop\_intent} & R & \intent{} identifier, reason & Retires an open \intent{} and revokes its worker's access. \\
\texttt{finish\_reason} & R & Description, outcome & Ends the review as planned or incomplete; acknowledges the reviewed revision. \\
\texttt{complete\_project} & R & Supporting \fact{}s, justification & Creates the completion \intent{} and marks the project complete. \\
\texttt{get\_intent\_status} & R & Optional \intent{} identifier & Reads current \intent{} state and published findings. \\
\texttt{get\_intent\_trajectory} & R & \intent{} identifier; optional request identifier & Reads model-request metadata or bounded input/output excerpts. \\
\texttt{reject\_task} & R/W & Explanation & Records that the execution declined its assignment. \\
\texttt{submission\_status} & R/W & None & Reads accepted identifiers and the execution's submission state. \\
\bottomrule
\end{tabular}
\endgroup

\end{table}

Creating an \intent{} requires a nonempty set of distinct, existing source \fact{} identifiers from the same project.
The protected \textsc{goal} is terminal, so source validation excludes it.
The server also enforces the configured limit on new \intent{}s per reasoning session.
Each accepted \intent{} becomes eligible for dispatch immediately, subject to available capacity.
Changing an \intent{}'s priority changes the ordering of pending work; a running worker retains its assignment.

A worker may publish several \fact{}s before concluding its \intent{}.
To finish, it must name a \fact{} submitted by that execution; the prompt requests a summary \fact{} when several findings need to be combined.
The reasoner can end a review with \texttt{finish\_reason}, recording either planned work or an unresolved outcome.
A planned outcome requires an open \intent{}.
When the reasoner judges the \textsc{goal} satisfied, \texttt{complete\_project} records the supporting \fact{}s and justification in a completion \intent{} whose outcome is \texttt{goal}, marks the project complete, and closes outstanding execution claims.
The benchmark's independent verifier evaluates the resulting answer.

\subsection{Ownership and transactional submissions}
\label{app:protocol-submissions}

Before starting an agent session, the dispatcher acquires an execution claim and obtains a credential bound to the project, role, worker identity, and assigned \intent{} when applicable.
The server checks these bindings, the active project state, and the claim's heartbeat lease for every new submission.
The dispatcher maintains exclusive ownership of an assigned \intent{} and permits at most one active reasoner per project.
A reasoning claim also records the reasoning-request revision under review.
Finishing that review acknowledges its input revision, preserving a later reasoning request if new evidence arrived during the session.

Each state-changing call carries a caller-generated \texttt{submission\_key} that identifies one logical operation within the execution.
The server checks the tool's role and arguments, applies the update, and stores its receipt in one database transaction.
An unsuccessful transaction leaves both the graph change and receipt uncommitted.
Scheduling notifications are published after the transaction commits, so a newly awakened dispatcher observes committed state.
The receipt provides the created identifiers and the result of the operation.
Agents use this confirmation to decide whether a submission succeeded.

When a reply is lost, retrying the submission with the same key and identical arguments retrieves the receipt without repeating the update.
Reusing a key with different arguments is rejected.
Receipt recovery remains available after normal task completion while the credential is retained and unrevoked.
Revoking a credential blocks further updates and receipt replay.

\subsection{Example graph records}
\label{app:protocol-example}

The following synthetic YAML excerpt shows how findings from two investigations support a subsequent \intent{}.
The reasoner passes \texttt{[f001, f002]} to \texttt{create\_intent}; the worker assigned to \texttt{i003} publishes \texttt{f003} and then passes that identifier to \texttt{finish\_intent}.
Descriptions are illustrative; metadata and earlier \intent{}s are omitted for readability.

\begin{quote}
\small
\begin{verbatim}
facts:
  - id: f001
    description: "Finding from the first investigation."
    source_intent_id: i001
  - id: f002
    description: "Finding from the second investigation."
    source_intent_id: i002
  - id: f003
    description: "Result obtained from the combined evidence."
    source_intent_id: i003
intents:
  - id: i003
    from: [f001, f002]
    to: f003
    description: "Investigate using both findings."
\end{verbatim}
\end{quote}

The \texttt{from} field identifies the supporting \fact{}s, while \texttt{source\_intent\_id} links each finding to the investigation that produced it.
Together, these fields record how the branches converge in the exported graph.

\subsection{Revision and cancellation}
\label{app:protocol-revision}

Evidence correction uses additional graph records.
When a finding is disputed, the reasoner can create a focused verification \intent{} and use the resulting \fact{}s to motivate revised work.
It can also call \texttt{drop\_intent} on an affected pending or running \intent{}, recording a concrete reason.
The server preserves the \intent{} and its published \fact{}s, marks the \intent{} dropped, clears its owner, and revokes the execution's access in the same transaction.
The dispatcher subsequently cancels the corresponding process.
A replacement investigation receives a fresh identifier, preserving the distinction between the retired direction and the revised one.

Earlier \fact{}s and completed descendants remain in the record.
The reasoner identifies affected work and cites corrective evidence when redirecting it; structural validation checks graph consistency, while evidence assessment remains part of the agent's investigation and review.

\clearpage
\section{Experimental Setup and Reproducibility}
\label{app:manifest}

\subsection{Task interfaces and verification}
\label{app:task-interfaces}
Both systems receive the same task information and operate under the same access permissions.
Answers are checked independently of the agents, which cannot access the grading references during solving.

\paragraph{FrontierMath.}
We evaluate both systems on the twelve public problems.
For each problem, the agent receives the statement and expected answer type, uses the container's Python interpreter and mathematical libraries, and submits its answer as a self-contained Python program.
The grader executes the submitted program in a fresh sandbox, serializes its return value, and checks that value in a separate verification step.
Variables and files from the solving process are unavailable in the grading sandbox.
Reference answers and checker code are accessible only to the evaluation controller.
We construct the local checkers from the public reference material and use them to verify submitted answers.

\paragraph{Cybench.}
Agents receive the original unguided complete-task prompt, permitted files, and target access, without benchmark subtask hints.
Each submission contains a list of answer strings, checked for exact matches against the task's final-answer criterion.
For tasks requiring several answers, each submission must contain the required set.
Where applicable, we check that the deployed target is consistent with the grading reference before solving begins.


\subsection{Agent configuration and prompt composition}
\label{app:experiment-prompts}
Both systems use Pi~0.85.1 with \texttt{glm-5.1-highspeed} through the model gateway.
GLM-5.1 was released on April~7, 2026~\citep{zai2026glm51}.
The gateway records model requests and provider-reported token usage.
\textsc{Direct} runs a single Pi session with its native system prompt, local tools, and context management in a fresh container without inherited state.
\system{} replaces the native system prompt with role-specific instructions and adds graph-based coordination tools while retaining Pi's local tools.
Its agents coordinate through shared graph state, with up to six execution slots shared by \textsc{reasoner} and \textsc{worker} sessions.

\paragraph{Submission instructions.}
\textsc{Direct} is instructed to solve the task and submit answers using a single agent loop, without delegation.
Agents in both systems are informed that incorrect submissions receive feedback and that the run ends upon acceptance or at the five-hour hard limit.
The hard cutoff prevents further submissions once the total solving budget, including any conclusion phase, expires.
For \system{}, the benchmark task and submission instructions form the project \textsc{origin}; the project \textsc{goal} is a receipt confirming acceptance by the independent grader.
We provide the role templates and their dynamic fields in Appendix~\ref{app:prompts}.

\paragraph{Execution environment.}
Each run uses a fresh, isolated project container based on the same benchmark-specific image for both systems.
\system{}'s \textsc{worker}s share the project container and filesystem.
The FrontierMath image includes Python with SymPy, NumPy, SciPy, and mpmath for symbolic and numerical computation, together with gmpy2, galois, pyadic, NetworkX, and Numba.
The Cybench image is based on Kali Linux and includes Nmap for network inspection, GDB and pwntools for binary analysis, and Z3 for constraint solving. For both systems, a restricted network policy disables public Internet access from the project containers.

\subsection{Deadlines and measurement}
\label{app:measurement}

The unified evaluation sets a five-hour hard limit for task solving. The hard cutoff closes submissions and requests termination; runs that reach the five-hour hard limit without an accepted answer are assigned a solving time of 300 minutes.

We measure solving time from agent activation to the first accepted submission or the five-hour hard cutoff.
This interval includes planning, tool use, coordination, and waiting.
We report total token usage by summing the available provider-reported token counts across all agent sessions within each run, following the provider's accounting convention.
\end{document}